\documentclass{article}

\usepackage{arxiv}

\usepackage[utf8]{inputenc} 
\usepackage[T1]{fontenc}    
\usepackage{hyperref}       
\usepackage{url}            
\usepackage{booktabs}       
\usepackage{amsfonts}       
\usepackage{nicefrac}       
\usepackage{microtype}      
\usepackage{lipsum}		
\usepackage{graphicx}
\usepackage{natbib}
\usepackage{doi}

\usepackage{xcolor}
\usepackage{caption}
\usepackage{subcaption}     
\usepackage{amsmath}
\usepackage{lscape}

\usepackage{siunitx}
\usepackage{booktabs}
\usepackage{graphicx}

\title{Machine Learning-Based Diabetes Detection Using Photoplethysmography Signal Features}

\author{Filipe A.~C.~Oliveira \\
	Heart Institute (InCor)\\
	Clinics Hospital University of \\
 Sao Paulo Medical School\\
	Sao Paulo - SP - Brazil \\
	\texttt{filipe.acoliveira@hc.fm.usp.br} \\
	\And
	Felipe M.~Dias \\
	Heart Institute (InCor)\\
	Clinics Hospital University of \\
 Sao Paulo Medical School\\
	Sao Paulo - SP - Brazil \\
	\texttt{f.dias@hc.fm.usp} \\
 \And
	Marcelo A.~F.~Toledo \\
	Heart Institute (InCor)\\
	Clinics Hospital University of \\
 Sao Paulo Medical School\\
	Sao Paulo - SP - Brazil \\
	\texttt{marcelo.arruda@hc.fm.usp.br} \\
 \And
	Diego A.~C.~Cardenas \\
	Heart Institute (InCor)\\
	Clinics Hospital University of \\
 Sao Paulo Medical School\\
	Sao Paulo - SP - Brazil \\
	\texttt{diego.cardona@hc.fm.usp.br} \\
 \And
    Douglas A.~Almeida \\
	Heart Institute (InCor)\\
	Clinics Hospital University of \\
 Sao Paulo Medical School\\
	Sao Paulo - SP - Brazil \\
	\texttt{douglas.andrade@hc.fm.usp.br} \\
 \And
	Estela ~Ribeiro \\
	Heart Institute (InCor)\\
	Clinics Hospital University of \\
 Sao Paulo Medical School\\
	Sao Paulo - SP - Brazil \\
	\texttt{estela.ribeiro@hc.fm.usp.br} \\
 \And
	Jose E.~Krieger \\
	Heart Institute (InCor)\\
	Clinics Hospital University of \\
 Sao Paulo Medical School\\
	Sao Paulo - SP - Brazil \\
	\texttt{j.krieger@hc.fm.usp.br} \\
  \And
	Marco A.~Gutierrez \\
	Heart Institute (InCor)\\
	Clinics Hospital University of \\
 Sao Paulo Medical School\\
	Sao Paulo - SP - Brazil \\
	\texttt{marco.gutierrez@incor.usp.br}
}

\date{}

\renewcommand{\shorttitle}{ML Diabetes Detection}

\hypersetup{
pdftitle={A template for the arxiv style},
pdfsubject={q-bio.NC, q-bio.QM},
pdfauthor={David S.~Hippocampus, Elias D.~Striatum},
pdfkeywords={First keyword, Second keyword, More},
}

\begin{document}
\maketitle

\begin{abstract}
Diabetes is a prevalent chronic condition that compromises the health of millions of people worldwide. Minimally invasive methods are needed to prevent and control diabetes but most devices for measuring glucose levels are invasive and not amenable for continuous monitoring. Here, we present an alternative method to overcome these shortcomings based on non-invasive optical photoplethysmography (PPG) for detecting diabetes. We classify non-Diabetic and Diabetic patients using the PPG signal and metadata for training Logistic Regression (LR) and eXtreme Gradient Boosting (XGBoost) algorithms. We used PPG signals from a publicly available dataset. To prevent overfitting, we divided the data into five folds for cross-validation. By ensuring that patients in the training set are not in the testing set, the model's performance can be evaluated on unseen subjects' data, providing a more accurate assessment of its generalization. 
Our model achieved an F1-Score and AUC of $58.8\pm20.0\%$ and $79.2\pm15.0\%$ for LR and $51.7\pm16.5\%$ and $73.6\pm17.0\%$ for XGBoost, respectively. Feature analysis suggested that PPG morphological features contains diabetes-related information alongside metadata. Our findings are within the same range reported in the literature, indicating that machine learning methods are promising for developing remote, non-invasive, and continuous measurement devices for detecting and preventing diabetes. 
\end{abstract}


\section{Introduction}

Diabetes is a chronic disease characterized by insufficient insulin release by the pancreas or lack of tissue response, leading to higher blood glucose levels (BGL) \cite{intro_defronzo}. When undetected or untreated, this higher glucose concentration may lead to several vascular illnesses, such as kidney and heart diseases. Diabetes management includes the prevention of such illnesses by constant BGL self-monitoring \cite{intro_kirk}.

Several devices for assessing the patient's BGL are either invasive based on fasting plasma glucose test, which requires a blood sample collection by a healthcare professional, or minimally invasive through finger prick blood test and continuous glucose monitors (CGM) for self-monitoring at home. However, all these methods require skin perforation, and only the CGM provides continuous measurements. Due to the high cost, discomfort, and risk provided by CGM, frequent BGL monitoring is not implemented at large. A non-invasive approach to detect diabetes would be valuable for the identification of early stages of diabetes in healthy and pre-Diabetic subjects, contributing to global efforts in preventing this disease and instrumental to assist monitor therapy interventions \cite{lamonte2005}. 

The Photoplethysmography (PPG) is one of the non-invasive approaches that has been studied for the detection of diabetes and pre-diabetes \cite{zanelli_review_2022}. 
The PPG is an optical technique for measuring changes in blood volume in the microvascular bed of tissues. This technique typically involves emitting light on the tissue and measuring changes in light absorption, reflection, or scattering to determine changes in blood volume \cite{MEJIAMEJIA2022}. Moreover, the quantification of certain molecules in the blood can be performed by employing light of particular molecules' peak absorption frequencies \cite{MUKKAMALA2022}. The canonical case is oximetry (SpO2). Besides that, PPG waveform can be used to measure other important cardiovascular and respiratory parameters, such as heart rate (HR), respiration rate, and blood pressure \cite{NITZAN2022}.

PPG can be applied as an indirect measure of a subject's hemodynamics information, conveying information pertaining to the condition of the blood vessel, such as the arterial stiffness, which is found to be elevated in individuals with diabetes \cite{pilt2013}.
 Some works suggest that Diabetic subjects have less prominent dicrotic notch \cite{Spigulis2002}, even though they used a reduced number of subjects in their analysis. 
 Other work show that the area under a PPG pulse is reduced when the level of HbA1c (glycated hemoglobin) level increase \cite{usman2011}. The HbA1c provides an estimation of a person's average blood glucose levels over the past two to three months and is used for diabetes diagnostic and management.

An advantage of PPG-based methods for assessing BGL is that they can provide continuous non-invasive measurements. These are not only important for monitoring, but also for detecting diabetes in its early stages of development. 
In \cite{reddy2017}, they applied an SVM method for predicting diabetes using features related to heart rate variability (HRV) and PPG morphology, using a fingertip pulse oximeter private dataset. 
In \cite{hettiarachchi_liang}, they proposed to predict diabetes based on morphological features extracted from PPG signals from a public dataset \cite{liang_dataset} using a Linear Discriminant Analysis method. 
In \cite{moreno2017}, they proposed a method for screening patients for diabetes by extracting features from the signal of a pulse oximeter. The features served as input for different classification algorithms, such as random forest and gradient boosting.
\cite{Srinivasan2021} made an analysis in the frequency domain using a convolutional neural network (CNN) with 30 seconds PPG scalograms and metadata over a large dataset.
More recently, \cite{zanelli2023} proposed a transfer learning approach to detect diabetes through 1-second raw PPG signals.

In this work, we presented a method to classify non-Diabetic and Diabetic patients using a set of features extracted from PPG signals and metadata for training machine learning models using Logistic Regression (LR) \cite{tolles2016} and eXtreme Gradient Boosting (XGBoost) \cite{chen2016} algorithms. We aimed to correlate the PPG signals morphological characteristics and the subject's diabetes status adjusted by the individual's metadata information. 
The method proposed in this paper utilizes shorter PPG signals, each segment lasting 2.1 seconds. This approach enables easy integration into various wearable devices, while also potentially offering benefits such as reduced computational costs and processing times.

\section{Materials and Methods}

\begin{figure}[!h]
	\centering
    \includegraphics[width=1\linewidth]{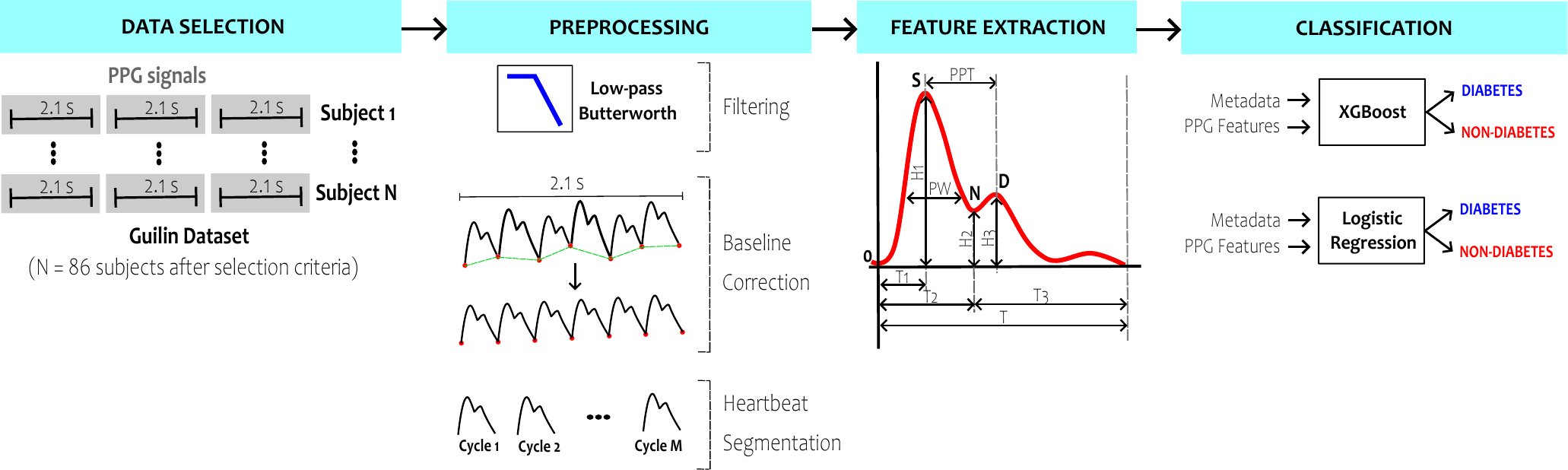}
	\caption{The proposed methodology for the classification of patients into Diabetes or non-Diabetes groups.}
	\label{fig: methodology}
\end{figure}

In this section, the description of the publicly available dataset used to classify the PPG signals into non-Diabetic or Diabetic patients is presented and also the data selection criteria. Next, we describe the proposed methodology for Diabetes classification, based on signal preprocessing and feature extraction of PPG signals, and the use of two algorithms for the classification step. 
Fig. \ref{fig: methodology} shows the proposed methodology for the classification between non-Diabetic and Diabetic groups.

\subsection{Data Selection}

We used a publicly available dataset provided by a study that contains physiological information and PPG data from 219 patients of Guilin People's Hospital in China \cite{liang_dataset}. The PPG signal was collected from the fingertip using the transmission method in infrared wavelength (905 nm) at a sample rate of 1 kHz. The PPG signal was collected from resting patients, where three 2.1-second PPG segments were captured from each patient.

The dataset contains information about three different classes of diseases: diabetes, hypertension, and cerebrovascular diseases, along with metadata information such as age, weight, height, body mass index (BMI), arterial blood pressure (ABP), and an unique patient ID number.

For this study, we divided the patients into two groups: non-Diabetic and Diabetic patients. We kept 59 healthy subjects in the non-Diabetic group by removing all patients with any stage of hypertension or cerebrovascular disease, and all 38 patients diagnosed with diabetes were kept in the other group. Due to segmentation errors and noisy signals, 11 subjects were excluded from the analysis. Table \ref{tab: dataset} show a description of the data used in this study.

\begin{table}[!h]
    \centering
    \caption{Summary of the data used.}
    \label{tab: dataset}
    \begin{tabular}{lcccccccc}
    \hline 
    \textbf{Class} & \textbf{ Subjects } & \textbf{ Male } & \textbf{ Cycles } & \textbf{ Age } & \textbf{ Height } & \textbf{ Weight } & \textbf{ HR } & \textbf{ BMI } \\ \hline
    \textit{Non-Diabetic} & $54$ & $22$ & $281$ & $45 \pm 16$ & $161 \pm 8$ & $56 \pm 11$ & $73 \pm 11$ & $22 \pm 4$ \\
    \textit{Diabetic} &     $32$ & $14$ & $172$ & $59 \pm 12$ & $160 \pm 8$ & $62 \pm 12$ & $74 \pm 11$ & $24 \pm 4$ \\ \hline
    \textit{Total} &        $86$ & $36$ & $453$ & $50 \pm 16$ & $161 \pm 8$ & $59 \pm 11$ & $74 \pm 11$ & $23 \pm 4$ \\ \hline
    \end{tabular}
\end{table}

\subsection{Preprocessing}

To remove high-frequency artifacts in the PPG, we preprocessed each signal segment with a 6\textsuperscript{th}-order Butterworth low-pass filter with a cutoff frequency of 16 Hz. During the acquisition of PPG signals, different artifacts may cause baseline oscillations. To address this issue, we applied the Fitting-based Sliding Window (FSW) algorithm instead of a high-pass-filter to remove the baseline, as the high-pass filter may eliminate important low-frequency characteristics present in the signal \cite{fsw_zhang}.

The FSW algorithm involves a sliding window that identifies the local minimums between each cycle (heartbeat) in the signal segments and uses them to fit the baseline. The fitted line was subtracted from the signal, removing the baseline fluctuations while preserving low-frequency characteristics. The valleys between the cycles were also used to segment the PPG cycles in each signal segment. The FSW algorithm details are described in \cite{fsw_zhang}.

We performed a visual inspection on the cycles extracted from PPG signals of all patients in the study. During this process, six Diabetic and five non-Diabetic patients were removed from the experiment due to segmentation flaws or presence of very noisy signals.

\subsection{Feature extraction}
Each PPG cycle segmented by the FSW served as input for the feature extraction algorithm. We used the same features presented in \cite{incor_ppg}, resulting in 104 features. The features were computed from each signal cycle (heartbeat) and its corresponding first and second derivatives. The features included a variety of values such as pulse width, area, intervals, peak-to-peak interval, and systolic amplitude. A detailed description of the features used here can be found in \cite{incor_ppg}, \cite{features_1}, \cite{features_2}, and \cite{features_3}.

The metadata (sex, age, height, weight, heart rate, and BMI) was used as features in the classification algorithm along with the PPG signal features. However, the data related to blood pressure was not included due to its strong relation to hypertensive and pre-hypertensive patients. This could create a potential bias in the classification algorithm as high blood pressure could be falsely associated with the presence of Diabetes, since most Diabetic patients included in the experiment have a degree of hypertension. 

This feature extraction step resulted in $453$ sets of features that were used to perform the detection of Diabetes on PPG signals, being $6$ features in each set related to the medadata information and $104$ related to features extracted from the PPG signal cycles.

\subsection{Classification}
We employed two well-known algorithms commonly used in machine learning classification tasks, the Logistic Regression  \cite{LR} and the XGBoost \cite{xgboost} algorithms. XGBoost is known for its high accuracy, ability to handle complex data, and scalability, but it's less interpretable. Logistic Regression, on the other hand, is a simpler model with better interpretability, but may not perform as well in complex datasets. 

$L_1$ penalty and \textit{LIBLINEAR} solver were used as hyperparameters for the LR algorithm.
For the XGBoost algorithm we used the tree booster with learning rate of $0.1$ and maximum tree depth of $30$. 
The weights for each class were adjusted according to the number of samples representing each class for both classifiers.

The performance of the LR and the XGBoost models was evaluated using statistical metrics: Accuracy ($Acc$), Sensitivity ($Se$), Specificity ($Sp$), F1-score, Positive Predictive Value ($PPV$), and the Area Under the ROC Curve ($AUC$).

The ``Diabetes'' column of the dataset was used as the label, considering one for Diabetics and zero for non-Diabetics. The classification was performed and evaluated using a 5-fold cross-validation strategy, which involves splitting the dataset into five equal parts or folds. In each iteration, four of the folds are used for training and one fold is used for testing. This process is repeated five times with each fold serving as the testing set once. The performance of the classifiers (LR and XGBoost) is then averaged across the five iterations. 

Features sets obtained from PPG cycles were split in each fold patient-wise. This way, features extracted from signals of the same patient are placed in training folds or testing fold, never in both. By allocating each patient in training and testing sets by their IDs, we avoided the classification bias caused by mixing information of the same patient in both training and testing folds in each iteration \cite{incor_ppg} \cite{deChazal}.

The overall performance of the classifiers is obtained by evaluating the mean and standard deviation of all statistical metrics obtained in each iteration. Moreover, we used the \textit{sklearn.inspection.permutation\_importance} algorithm to determine feature importance for both classifiers. Since we have a 5-fold strategy, the mean feature importance for all 5-folds was computed.

\section{Results}

This study used data from 86 subjects, with 453 cycles extracted using the FSW method. Out of these cycles, 172 belong to Diabetic patients. A total of 110 features were used as input for the classifiers, with 104 of these being morphological features extracted from each PPG cycle and the remaining 6 features comprising metadata of the patients. 

The goal of the study was to classify non-Diabetic and Diabetic patients by using LR and XGBoost classifiers on features extracted from PPG cycles. The 5-fold cross-validation strategy permitted the evaluation of the classifiers' performance on multiple iterations and reduced the impact of any random fluctuations in the data. 

\begin{figure}[!h]
        
      \centering
      \includegraphics[width=0.6\linewidth]{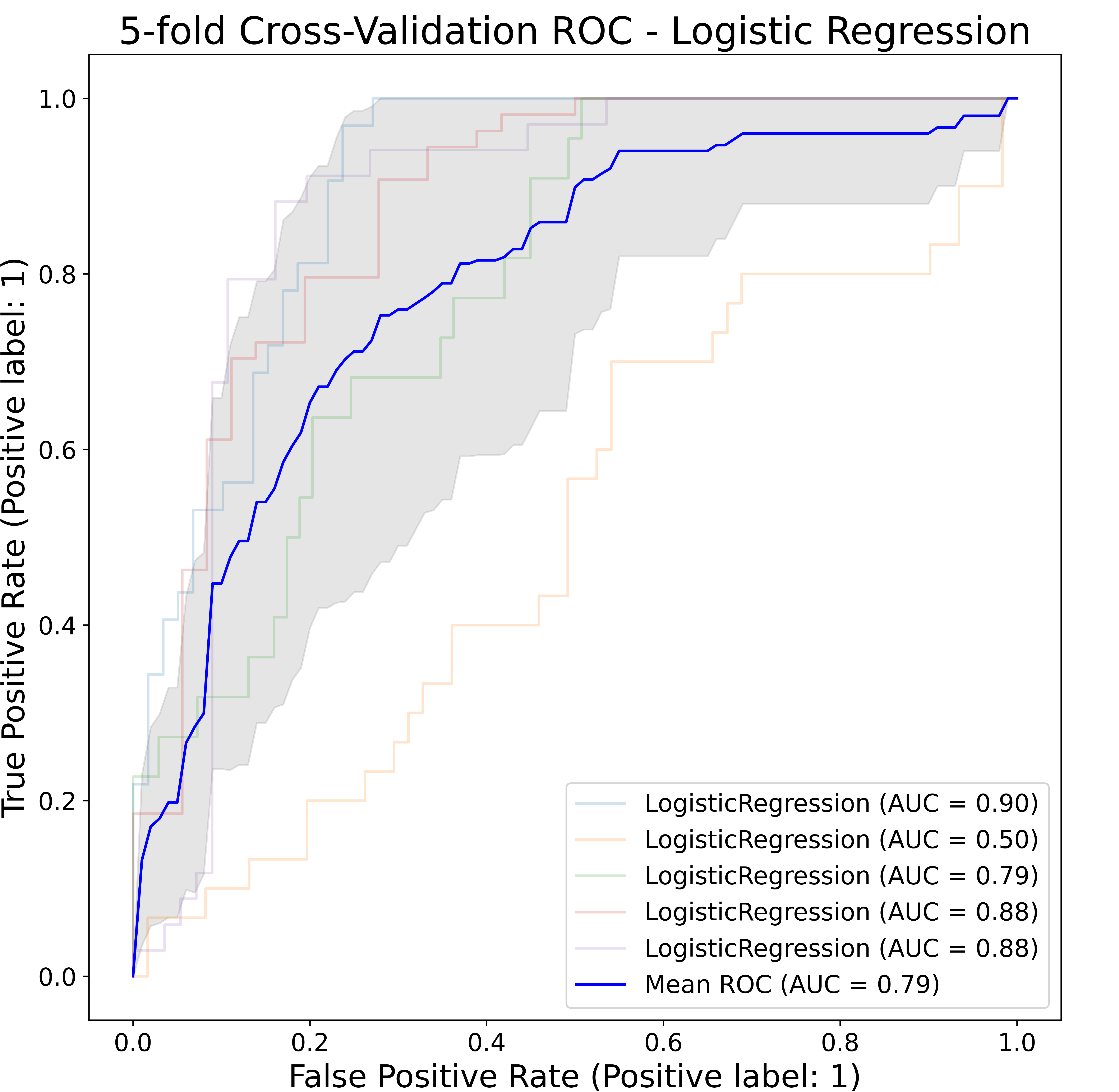}
      \caption{Receiver Operating Curve (ROC) and AUC for each fold of the Logistic Regression method.}
      \label{fig: res_roc_RL}
\end{figure}

\begin{figure}[!h]
      \centering
      \includegraphics[width=0.6\linewidth]{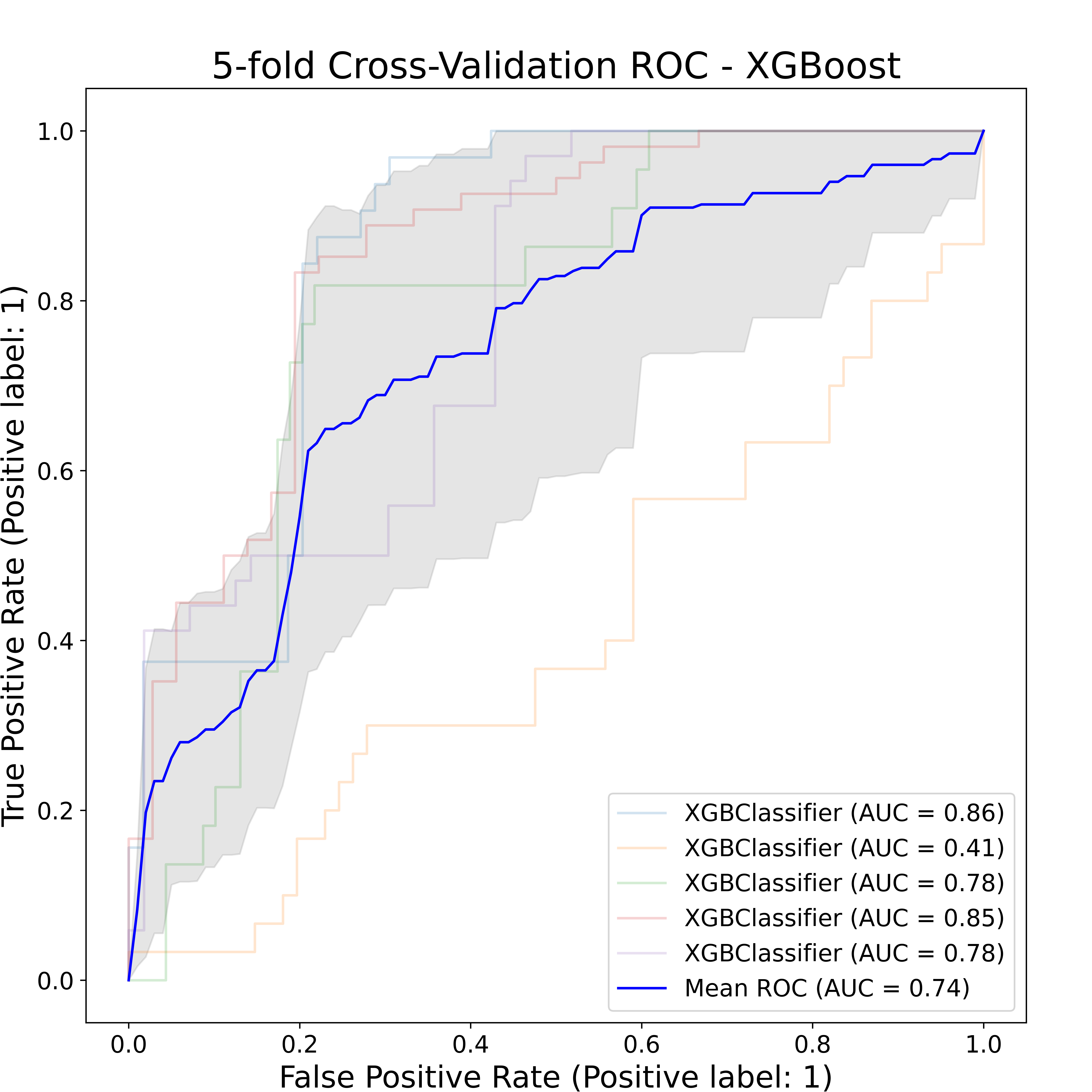}
      \caption{Receiver Operating Curve (ROC) and AUC for each fold of the XGBoost method.}
      \label{fig: res_roc_Xg}
\end{figure}

Table \ref{tab: results} summarizes the overall performance of the proposed methods for the binary classification between Diabetic/non-Diabetic. The metrics are presented as the mean and standard deviation of the five iterations of the cross-validation strategy. In addition, Figs. \ref{fig: res_roc_RL} and \ref{fig: res_roc_Xg} show the AUC for the LR and XGBoost methods, respectively.

\begin{table}[!h]
    \centering
    \caption{Comparison of the performance results of our two proposed algorithms for Diabetic/Non-Diabetic classification.}
    \label{tab: results}
    \begin{tabular}{lcc}
        \hline
        \textbf{Metrics  } & \textbf{ Logistic Regression } & \textbf{ XGBoost } \\ \hline
        \textit{Se} & $66.4 \pm 26.6\%$ & $56.3 \pm 22.1\%$ \\
        \textit{Spe} &  $75.1 \pm 18.9\%$ & $70.5 \pm 16.0\%$ \\
        \textit{F1-score} & $58.9 \pm 20.0\%$ & $51.7 \pm 16.5\%$ \\
        \textit{Acc} & $70.0 \pm 12.2\%$ & $64.5 \pm 9.8\%$ \\
        \textit{PPV} & $61.1 \pm 24.1\%$ &  $54.0 \pm 21.3\%$ \\
        \textit{AUC} & $79.2 \pm 15.0\%$ & $73.6 \pm 17.0\%$ \\ \hline
    \end{tabular}
\end{table}

Table \ref{tab: results} shows the classification metrics of this study, which achieved F1-Score and AUC of $58.8\pm20.0\%$ and $79.2\pm15.0\%$ for LR and $51.7\pm16.5\%$ and $73.6\pm17.0\%$ for XGBoost, respectively. The best achieved AUC with LR model is comparable to the AUC of $79\pm15\%$, reported in \cite{hettiarachchi_liang}, which used the same dataset for diabetes detection.

\begin{figure}[!h]
      \centering
      \includegraphics[width=0.8\linewidth]{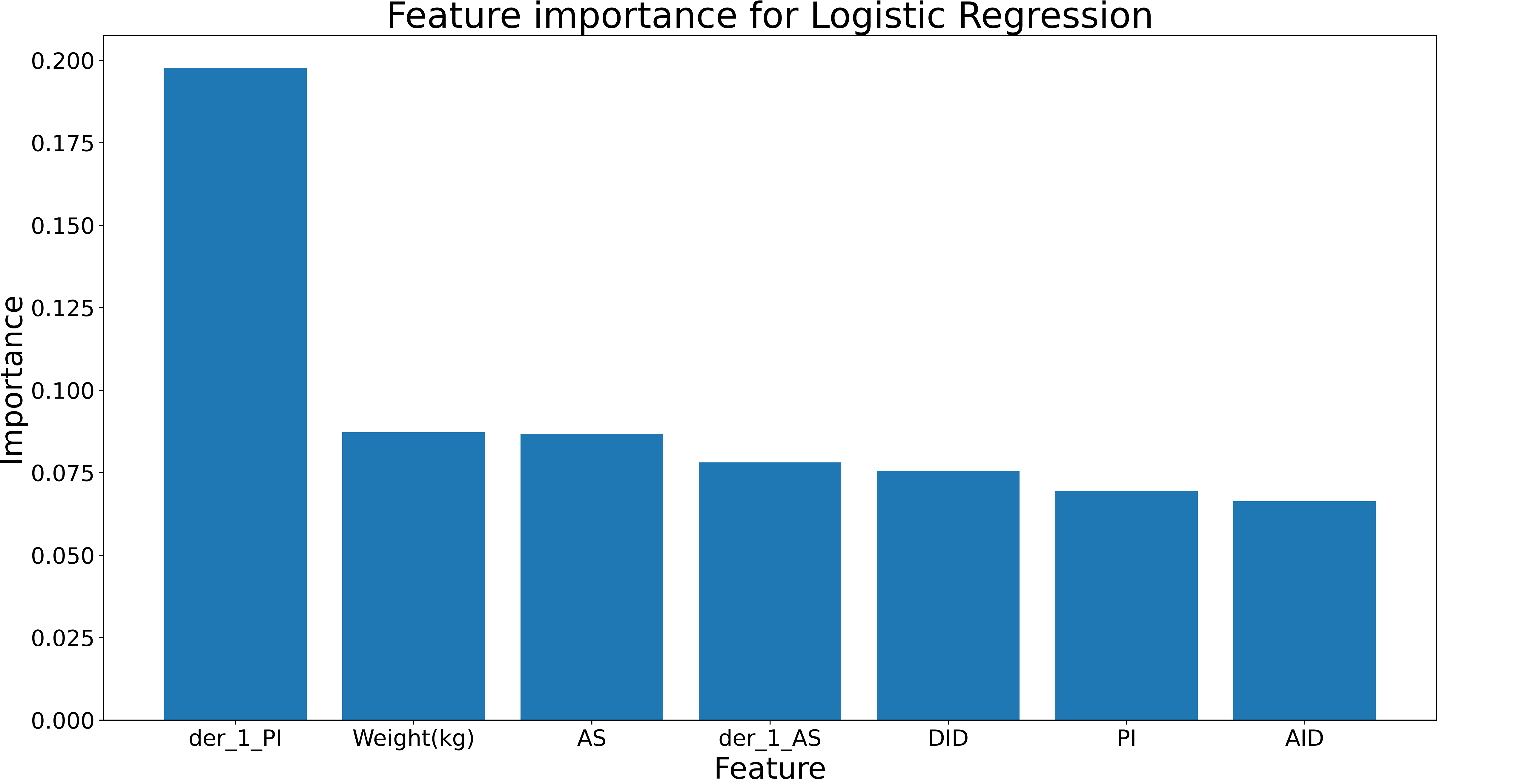}
      \caption{Importance of the 7 most significant features for Logistic Regression. The values represent the the mean of the five folds.}
      \label{fig: feat_importance_lr}
\end{figure}

\begin{figure}[!h]
      \centering
      \includegraphics[width=0.8\linewidth]{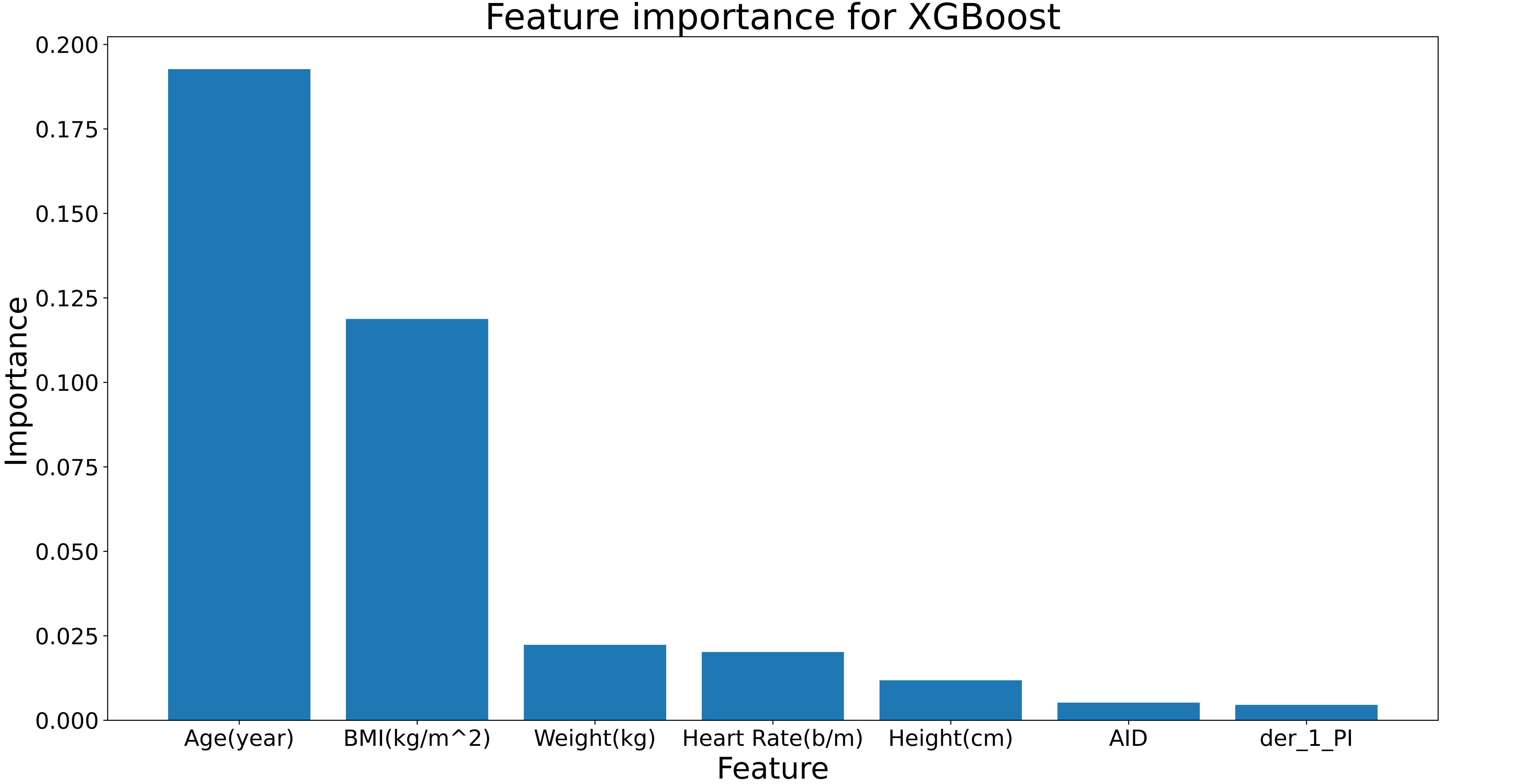}
      \caption{Importance of the 7 most significant features for XGBoost. The values represent the the mean of the five folds.}
      \label{fig: feat_importance_xgboost}
\end{figure}

Additionally, we present the feature importance measures, displaying the 7 most significant features for the two proposed models' ability to predict the outcomes. The importance and name of each feature is shown in Fig. \ref{fig: feat_importance_lr} for Logistic Regression and in Fig. \ref{fig: feat_importance_xgboost} for XGBoost. 
Based on the feature importance results obtained from both algorithms, we have compiled a summary of the six most significant PPG features in Table \ref{tab: feat_importance}. We have excluded from this table the features associated with metadata, namely age, weight, height, and BMI, as they are self-explanatory.

\begin{table*}[!h]
\renewcommand{\arraystretch}{1.2}
    \centering
    \caption{Description of the six most significant PPG signal features of the experiment.}
    \label{tab: feat_importance}
    \begin{tabular}{ p{0.2\textwidth}p{0.7\textwidth} }
        \hline
        \textbf{ Name } & \textbf{ Description } \\ \hline
        
        \textit{der\_1\_PI} & Intensity of the 1st inflection point in the 1st derivative of the PPG cycle \\
        \textit{AS} & Slope of the systolic portion in the PPG cycle (from the beginning of the cycle to the systolic peak) \\
        
        \textit{der\_1\_AS} &  Slope of the line up to the 1st inflection point (from the beginning of the 1st derivative to the 1st inflection point) \\
        
        \textit{DID} & Difference in intensity between the systolic peak and the end of the PPG cycle \\
        
        \textit{AID} & Difference in intensity between the beginning of the PPG cycle and the systolic peak \\
        
        \textit{PI} &  Intensity of the systolic peak in the PPG cycle normalized by the average value (baseline) \\ \hline
    \end{tabular}
\end{table*}

\section{Discussion}

In this study, we proposed a method aimed at aiding in the diagnosis of diabetes based on PPG signals. This is an exploratory study that correlates the morphological features extracted from PPG signals with metadata information from the subjects. 

We show that it is possible to achieve results comparable to the state-of-the-art using a limited dataset, with features extracted from single heartbeats (cycles) of the PPG signals. Table \ref{tab: literature_comparison} displays a comparison of works that propose to detect diabetes through PPG signals. 
\cite{hettiarachchi_liang} used the same dataset as our study; however, they used features extracted from the whole 2.1 s segment available, reducing the input size of their models. 
\cite{moreno2017}, on the other hand, used a private dataset with one-minute long segments registered from 1,170 subjects and achieved results in the same range as ours (AUC 69.4\%, Se 65\% Spe 64\%) using different machine learning techniques. 
\cite{zanelli2023}, the most recent work on this topic, used 1 s segments of PPG raw signals from 100 subjects to predict diabetes. Even though they proposed a transfer learning strategy, heir best-performing model was the one without transfer learning, combining age and sex metadata information.

Different from the works on Table \ref{tab: literature_comparison}, \cite{NIRALA2019} achieved Acc 97.87\%, Se 98.78\% and Spe 96.61\%, using data from 141 subjects. However, they randomly partitioned the data into k-equal folds on a k-fold strategy, implicating that they mixed the same subject data on the training and test partitions. \cite{gupta2022} also did the same approach, not clearly specifying the dataset division. Previous research has shown that utilizing the same subject for both the training and test sets can yield better outcomes, as data leakage may occur due to the high interdependence among intra-subject heartbeats \cite{deChazal, incor_ppg}. \cite{reddy2017}, on the other hand, achieved Acc 89\%, Se 90\% and Spe 88\%, using data from 100 subjects. Nevertheless, they didn't make it clear if during partition of the data on the k-fold strategy they ensure that the same subject signals wasn't on both training and test sets.

\begin{table}[!h]
\centering
\caption{Comparison of the performance results of our two proposed algorithms for Diabetes detection through PPG signal and the related state-of-the-art works.}
\label{tab: literature_comparison}
\resizebox{\textwidth}{!}{%
\renewcommand{\arraystretch}{2}
\begin{tabular}{cccccccc}
\toprule
\textbf{Reference} & \textbf{F1-score} & \textbf{AUC} & \textbf{Accuracy} & \textbf{Sensitivity} & \textbf{Specificity} & \textbf{Dataset} & \textbf{Method} \\
\midrule
\cite{moreno2017} & -- & 70.0 & -- & 80.0 & 48.0 & Private (s = 1,170) & RF and GB - PPG features and metadata \\
\cite{reddy2017} & -- & -- & 89.0 & 90.0 & 88.0 & Private (s = 100) & SVM - PPG features \\
\cite{hettiarachchi_liang} & 71.0 $\pm$ 15.0 & 79.0 $\pm$ 15.0 & 71.0 & -- & -- & \cite{liang_dataset} (s = 64) & LDA - PPG features and metadata \\
\cite{hettiarachchi_liang} & 69.0 $\pm$ 10.0 & 74.0 $\pm$ 17.0 & 79.0 & -- & -- & \cite{liang_dataset} (s = 64) & SVM - PPG features and metadata \\
\cite{NIRALA2019} & 98.18 & 97.69 & 97.87 & 98.78 & 96.61 & Private (s = 141) & SVM - 37 PPG features \\
\cite{avram2020} & -- & 76.6 & -- & 75.0 & 65.4 & Private (s = 55,433) & CNN - PPG raw signal and metadata \\
\cite{Srinivasan2021} & -- & 83.0 & 76.34 & 76.66 & 76.11 & MIMIC III (s = 808) & 2D CNN - PPG scalogram and metadata \\
\cite{gupta2022} & 99.0 & -- & 98.52 & 99.0 & 96.0 & \cite{liang_dataset} (s = 219) & RF - PPG features \\
\cite{zanelli2023} & 46.15 & 56.50 & -- & 75.0 & 76.0 & Private (s = 100) & CNN - PPG raw signal, age and sex \\
\cite{zanelli2023} & 30.77 & 61.0 & -- & 50.0 & 72.0 & Private (s = 100) & CNN and Transfer Learning - PPG raw signal, age and sex \\
Our method: XGBoost & 51.7 $\pm$ 16.5 & 73.6 $\pm$ 17.0 & 64.5 $\pm$ 9.8 & 56.3 $\pm$ 22.1 & 70.5 $\pm$ 16.0 & \cite{liang_dataset} (s = 86) & XGBoost - 104 PPG features and 6 metadata \\
Our method: LR & 58.9 $\pm$ 20.0 & 79.2 $\pm$ 15.0 & 70.0 $\pm$ 12.2 & 66.4 $\pm$ 26.6 & 75.1 $\pm$ 18.9 & \cite{liang_dataset} (s = 86) & LR - 104 PPG features and 6 metadata \\
\bottomrule
\end{tabular}%
}
\end{table}

\begin{figure}[!h]
      \centering
      \includegraphics[width=1\linewidth]{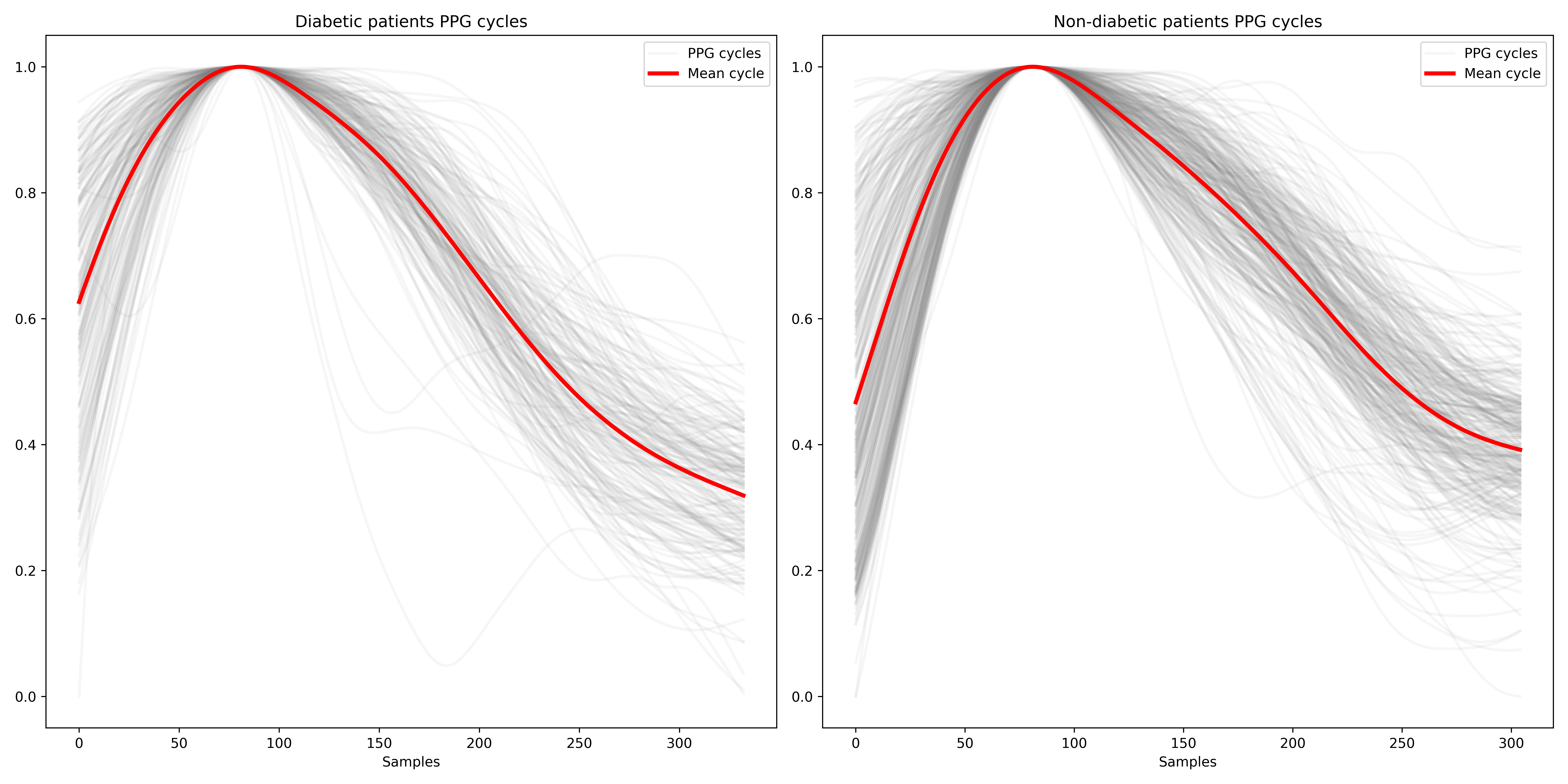}
      \caption{Average heartbeat (red) for Diabetic and non-Diabetic groups. All PPG cycles were normalized and aligned by the highest peak in the cycle.}
      \label{fig: mean_cycle}
\end{figure}

\cite{NIRALA2019} suggest in their work that the absence of a dicrotic notch on PPG waves can be observed in Diabetic subjects. \cite{zanelli2023} also compares a waveform from a Diabetic and a non-Diabetic subjects, highlighting the absence of a dicrotic notch in the former. However, they imply that these differences are not always readily apparent. On the contrary, \cite{Srinivasan2021} indicates that despite using a much larger dataset than the aforementioned studies, it is not possible to observe this dicrotic notch difference between the two groups. Our work aligns with Srinivasan's findings. In Figure \ref{fig: mean_cycle}, we present the average heartbeat of both Diabetic and non-Diabetic subjects in our dataset, demonstrating no discernible difference in regarding the presence of dicrotic notch. Moreover, the absence of dicrotic notch may be related to aging or other factors \cite{dicrota}.

\cite{Srinivasan2021} suggest that their work outperformed the studies based on PPG features due to this similarity of PPG wave morphology. Even though \cite{zanelli2023} also used a deep learning approach, they used a much smaller input signal (1 s vs 30 s length). Moreover, Srinivasan's study implicate that frequency domain information may be important for diabetes detection. Future works should be done in this regard.

Heart Rate Variability (HRV) is broadly used as a feature in diabetes detection and BGL estimation studies \cite{reddy2017} \cite{chu2021}. However, this feature may not be calculated with confidence in short segments of data such as the available in our dataset. For this reason, the HRV was not estimated for this experiment.

The small number of patients in the dataset raises concerns about potential bias in the classifier's performance. There is a possibility that the classifier may learn to estimate other clinical parameters, such as blood pressure or other clinical conditions, instead of focusing solely on Diabetes classification. In addition, several studies involving Diabetes classification do not explicitly consider the bias caused by mixing PPG signals from the same patient in the training and testing steps \cite{panwar2020}, which may result in better metrics due to overfitting. We addressed this issue by adopting the 5-fold cross-validation strategy considering the patient's identification number while splitting the data.

Comparing the significance of each feature shown in Figs. \ref{fig: feat_importance_lr} and \ref{fig: feat_importance_xgboost}, it is seen that the Logistc Regression algorithm considered mostly the PPG-related features as important for predicting diabetes. On the other hand, the XGBoost algorithm made predictions considering almost exclusively the patient's metadata information. The better results provided by LR algorithm suggest that the PPG cycles may indeed carry diabetes-related information. Both classification algorithms consider as important two specific PPG features: \textit{der\_1\_PI} and \textit{AID}. These features use information from the systolic wave between its onset and peak, indicating a potential value of this portion of the PPG wave for further analysis.

One of the reasons for the lack of studies using this dataset may be related to its reduced number of patients and its short time signals, considering that each patient has only six seconds of PPG signal divided into three non-continuous segments. The visual analysis shows that most of the PPG segments display incomplete cycles at the signal edges, which led to the disposal of large portions of PPG data. This aspect may impair the performance of segmentation algorithms, such as the FSW. Such impairment seen in this work led to the exclusion of eleven patients from the experiment due to segmentation errors and noisy signals. The FSW algorithm was first introduced in \cite{fsw_zhang} for removing the baseline and segmenting PPG signals. However, when applied to signals with a small number of cycles, it tends to discard the edges, resulting in a reduction of the data passed to the feature extraction algorithm and, consequently, a lower number of computed features, which negatively affects the performance and generalization of the classifiers.

The clinical use of PPG-based diabetes detection algorithms demands higher accuracy and reliability on different patients with varying medical conditions. Therefore, in order to enhance the effectiveness and reliability of the findings, ensuring their generalizability and applicability, future studies on the detection of diabetes using wearable devices should be conducted on a larger and more diverse dataset, including patients with different medical conditions, ages, and ethnicities. 
Thus, we urge for the need of publicly available well-annotated datasets on this theme in order to improve the effectiveness of PPG-based diabetes detection algorithms and facilitate its translation to clinical practice.

\section{Conclusion}

Our study proposed a method for diagnosing diabetes based on PPG signals, utilizing morphological features extracted from single beats of the PPG signals. We achieved comparable results to state-of-the-art studies using a limited dataset. Comparison with other studies showed that our approach yielded promising accuracy, sensitivity, and specificity values. However, the small number of patients in our dataset and the short duration of the PPG signals available for the analysis may have impacted our results. Additionally, publicly available well-annotated datasets in this field are needed for further advance research in this area.

\bibliographystyle{unsrtnat}
\bibliography{references}  






\end{document}